\documentclass[11pt,a4paper]{article}
\usepackage[hyperref]{emnlp2020}
\usepackage{times}
\usepackage{latexsym}

\makeatletter
\def\blfootnote{\xdef\@thefnmark{}\@footnotetext}
\makeatother

\usepackage{microtype}

\usepackage{mystyle}
\usepackage{subcaption} 

\usepackage{tikz}%

\usepackage{algorithm}
\usepackage[noend]{algpseudocode}

\usepackage{pgfplots}

\aclfinalcopy 
\def\aclpaperid{3272} 

\title{Scaling Hidden Markov Language Models}

\author{Justin T. Chiu {\normalfont and} Alexander M. Rush\\
  Department of Computer Science \\
  Cornell Tech \\
  \texttt{\{jtc257,arush\}@cornell.edu}\\}

\date{}

\begin{document}
\maketitle
\begin{abstract}
The hidden Markov model (HMM) is a fundamental tool for sequence modeling that 
cleanly separates the hidden state from the emission structure.
However, this separation makes it difficult to fit HMMs to large datasets in modern NLP, 
and they have fallen out of use due to very poor performance 
compared to fully observed models. This work revisits the challenge of 
scaling HMMs to language modeling datasets,
taking ideas from recent approaches to neural modeling.
We propose methods for scaling HMMs to massive state spaces
while maintaining efficient exact inference, a compact parameterization,
and effective regularization.
Experiments show that this approach leads to models that are more accurate
than previous HMM and n-gram-based methods,
making progress towards the performance of state-of-the-art neural models.
\end{abstract}

\blfootnote{Code available at \href{https://github.com/harvardnlp/hmm-lm}{github.com/harvardnlp/hmm-lm}}
\section{Introduction}
Hidden Markov models (HMMs) are a fundamental latent-variable model for sequential data,
with a rich history in NLP.
They have been used extensively in tasks such as
tagging \citep{merialdo1994tagging}, alignment \citep{vogel1996hmm},
and even, in a few cases, language modeling \citep{kuhn1994hmmlm,huang2011thesis}. 
Compared to other sequence models, HMMs are appealing since they 
fully separate the process of generating hidden states from observations,
while allowing for exact posterior inference. 

State-of-the-art systems in NLP have moved away from utilizing latent hidden states
and toward deterministic deep neural models.
We take several lessons from the success of neural models for NLP tasks:
(a) model size is critical for accuracy,
e.g. large LSTMs \cite{zaremba2014lstm} show marked improvements in performance;
(b) the right parameterization is critically important for representation learning,
e.g. a feedforward model \cite{bengio2003nlm}
can have the same distributional assumptions as an n-gram model while performing significantly better;
(c) dropout is key to achieving strong performance \citep{zaremba2014lstm,merity2017awdlstm}.

We revisit HMMs for language modeling
as an alternative to modern neural models,
while considering key empirical lessons from these approaches. 
Towards that goal, we introduce three techniques:
a modeling constraint that allows us to use a large number of hidden states 
while maintaining efficient exact inference,
a neural parameterization that improves
generalization while remaining faithful to the
probabilistic structure of the HMM,
and a variant of dropout that both improves accuracy
and halves the computational overhead during training.

Experiments employ HMMs on two language modeling datasets.
Our approach allows us to train an HMM with tens of thousands of states
while maintaining efficiency and
significantly outperforming past HMMs as well as n-gram models.

\section{Related Work}
\label{sec:rw}
In order to improve the performance of HMMs on language modeling,
several recent papers have combined HMMs with neural networks.
\citet{buys2018hmm} develop an approach to relax HMMs,
but their models either perform poorly or alter the probabilistic structure to resemble an RNN. 
\citet{krakovna2016hmm} utilize model combination with an RNN to connect both approaches in a
small state-space model.
Our method instead focuses on scaling pure HMMs to a large number of states.

Prior work has also considered neural parameterizations of HMMs. 
\citet{tran2016hmm} demonstrate improvements in POS induction with a
neural parameterization of an HMM.
They consider small state spaces,
as the goal is tag induction rather than language modeling.\footnote{
Other work has used neural parameterization for structured models, such as 
dependency models \citep{han2017dependency},
hidden semi-Markov models \citep{wiseman2018hsmm},
and context free grammars \citep{kim2019cpcfg}.
}

Most similar to this work are the large HMM models of 
\citet{dedieu2019learning}. They introduce a sparsity constraint
in order to train a 30K state non-neural HMM for character-level language modeling;
however, their constraint precludes application to large vocabularies.
We overcome this limitation and train models with 
neural parameterizations on word-level language modeling.

Finally, another approach for scaling state spaces is to
grow from small to big via a split-merge process
\citep{petrov2006splitmerge,huang2011thesis}.
In particular, \citet{huang2011thesis} learn an HMM for language modeling
via this process.
As fixed-size state spaces are amenable to batching on modern hardware,
we leave split-merge procedures for future work. 

\section{Background: HMMs}

We are interested in learning a distribution over observed tokens
$\bx = \langle x_1, \ldots, x_T \rangle$, with each token $x_t$
an element of the finite vocabulary $\mcX$.
Hidden Markov models (HMMs) specify a joint distribution over 
observed tokens $\bx$ and discrete latent states $\bz = \langle z_1, \ldots, z_T \rangle$,
with each $z_t$ from the finite set $\mcZ$.
For notational convenience, we define the starting state $z_0=\epsilon$.
This yields the joint distribution,
\begin{equation}
p(\bx, \bz; \theta)
= \prod_{t=1}^T p(x_t\mid z_t)p(z_t \mid z_{t-1}).
\end{equation}
\noindent We refer to the transition and emission matrices as the distributional parameters of the HMM.
Specifically, let $\bA \in[0,1]^{|\mcZ|\times|\mcZ|}$ be the transition probabilities and
$\bO \in[0,1]^{|\mcZ|\times|\mcX|}$ the emission probabilities,
\begin{equation}
p(z_t \mid z_{t-1}) = A_{z_{t-1}z_t}  \quad
p(x_t \mid z_t) = O_{z_tx_t}.
\end{equation}

We distinguish between two types of model parameterizations: \textit{scalar} and \textit{neural},
where the model parameters are given by $\theta$.
A scalar parameterization sets the model parameters equal to the distributional
parameters, so that $\theta = \set{\bA, \bO}$,
resulting in $O(|\mcZ|^2 + |\mcZ||\mcX|)$ model parameters.
A neural parameterization instead generates the distributional parameters
from a neural network (with parameters $\theta$),
decoupling the size of $\theta$ from $\bA,\bO$.
This decoupling gives us the ability to choose
between compact or overparameterized $\theta$ (relative to $\bA,\bO$).
As we scale to large state spaces,
we take advantage of compact neural parameterizations.

In order to fit an HMM to data $\bx$,
we must marginalize over the latent states to obtain the likelihood
$p(\bx) = \sum_{\bz}p(\bx,\bz)$.
This sum can be computed in time $O(T|\mcZ|^2)$ via the forward algorithm,
which becomes prohibitive if the number of latent states $|\mcZ|$ is large.
We can then optimize the likelihood 
with gradient ascent (or alternative variants of expectation maximization).

\noindent \textbf{HMMs and RNNs}
Although the forward algorithm resembles that of the forward pass in a recurrent neural network (RNN)
\citep{buys2018hmm}, there are key representational differences.
RNNs do not decouple the latent dynamics from the observed.
This often leads to improved accuracy,
but precludes posterior inference which is useful for interpretability.
A further benefit of HMMs over RNNs is that
their associative structure allows for parallel inference
via the prefix-sum algorithm \cite{ladner1980prefix}.\footnote{
Quasi-RNNs \citep{bradbury2016qrnn} also have a (parallel) logarithmic dependency on $T$
by applying the same prefix-sum trick, but do not model uncertainty over
latent dynamics.}
Finally, HMMs bottleneck information from every timestep through a discrete hidden state. 
NLP has a long history of utilizing discrete representations,
and discrete representations may yield interesting results.
For example, recent work has found that discrete latent variables
work well in low-resource regimes \citep{jin2020discrete}.

\section{Scaling HMMs}
\label{sec:methods}

We propose three extensions to scale HMMs for better language modeling performance:
blocked emissions, which allow for very large models;
neural parameterization, which makes it easy for states to share model parameters;
and state dropout, which encourages broader state usage. 

\begin{figure}[t]
\centering
\includegraphics[height=1.7in]{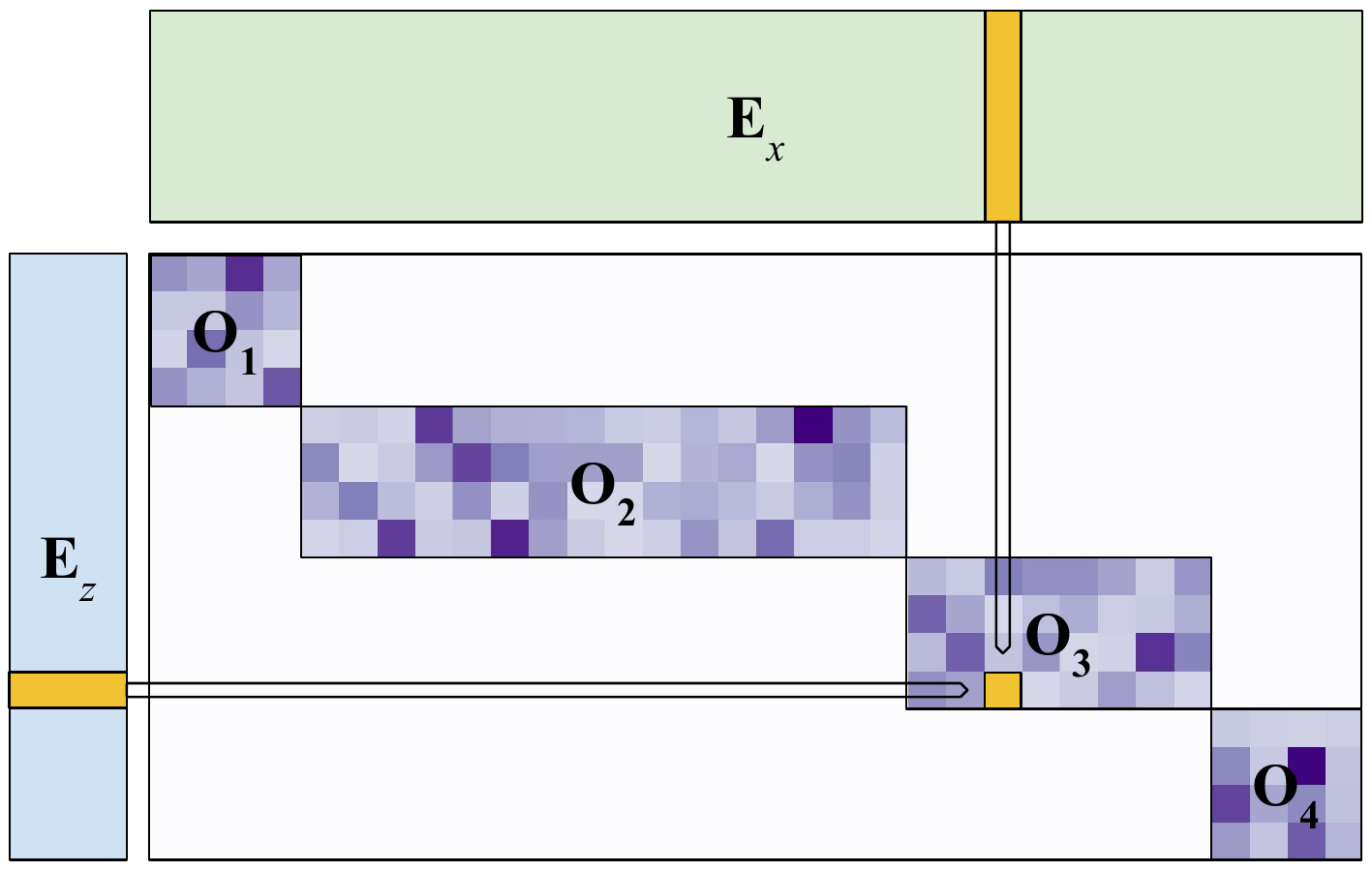}
\caption{\label{fig:emit}
The emission matrix as a set of blocks $\mathbf{O}_1, \ldots, \mathbf{O}_4$
with fixed number of states $k$.
The columns of each block may vary, as there is no constraint on the number of words
a state can emit.
Each non-zero cell is constructed from an MLP applied to word $\mathbf{E}_x$
and state $\mathbf{E}_z$ embeddings.
}
\end{figure}

\vspace{0.2cm}

\noindent
\textbf{Blocked Emissions}
Our main goal is to apply a HMM with a large number of hidden states to 
learn the underlying dynamics of language data.
However, the $O(T|\mcZ|^2)$ complexity of marginal inference
practically limits the number of HMM states.
We can get around this limit by making an assumption on the HMM emission matrix $\mathbf{O}$.
As noted by \citet{dedieu2019learning}, 
restricting the number of states that can produce each word can improve inference complexity. 
We utilize a slightly stronger assumption on the model:
a) states are partitioned into $M$ equal sized groups each
of which emit the same subset of words, 
and b) each word is only admitted by one group of
$k= |\mcZ| / M$ states which we indicate as $\mcZ_x \subset {\cal Z}$. 

We implement this group structure through a set of blocked emissions,
each corresponding to one of the $M$ state groups,
\begin{equation*}
\mathbf{O} = \begin{bmatrix} \mathbf{O}_1 & 0 & 0 \\ 0 & $\dots$ & 0 \\ 0 & 0 & \mathbf{O}_M \\
\end{bmatrix}
\end{equation*}
where $\mathbf{O}_m \in \mathbb{R}^{ k \times |\mcX_m|}$.
Figure~\ref{fig:emit} shows these emission blocks.
Each block matrix $\mathbf{O}_m$ gives the probabilities for emitting tokens $\mcX_m$ for states
in group $m$, i.e. states $(m-1)k$ through $mk$.

With this constraint, exact marginalization can be computed via 
\begin{equation}
\label{eqn:sparse_marginalization}
\begin{aligned}
p(\bx) &= \sum_{z_1 \in \mcZ_{x_1}} p(z_1\mid z_0)p(x_1 \mid z_1) \times\\
    &\cdots
    \sum_{z_T \in \mcZ_{x_T}} p(z_T \mid z_{T-1})p(x_T \mid z_T)
\end{aligned}
\end{equation}
Since there are only $k$ states with nonzero probability of occurring
at every timestep, we only need to consider transitioning from the $|\mcZ_{x_t}| = k$
previous states to the next $|\mcZ_{x_{t+1}}|=k$ states,
resulting in $O(k^2)$ operations per timestep.
This gives a serial complexity of $O(Tk^2)$.\footnote{
This can be sped up on a parallel machine to $O(\log(T)k^2)$
via a binary reduction.
}

\vspace{0.2cm}

\noindent
\textbf{Neural Parameterization}
A larger state space allows for longer HMM memory, but it also may require 
more parameters.
Even with blocked emissions, the scalar model parameterization of an HMM grows
as $O(|\mcZ|^2)$ due to the transition matrix. A neural parameterization allows us to 
share parameters between words and states to capture common structure. 

Our parameterization uses an embedding for each state in
$\mcZ$ ($\mathbf{E}_z \in \mathbb{R}^{|\mcZ| \times h}$)
and each token in $\mcX$ ($\mathbf{E}_x \in \mathbb{R}^{|\mcX| \times h}$).
From these we can create representations for leaving and entering a state,
as well as emitting a word: 
\[ \mathbf{H}_{\textrm{out}},\mathbf{H}_{\textrm{in}},\mathbf{H}_\textrm{emit}
 = \text{MLP}( \mathbf{E}_z ) \] 
with all in $\mathbb{R}^{|\mcZ|\times h}$.
The HMM distributional parameters are then computed as,\footnote{
As an optimization, one could only compute the nonzero emission matrix blocks saving space and time.
In practice we compute the full matrix as in the equation.
} 
\begin{equation}
\begin{aligned}
\mathbf{O} \propto \exp (\mathbf{H}_\textrm{emit}\mathbf{E}_x ^\top) \qquad
\mathbf{A} \propto \exp (\mathbf{H}_\textrm{in}\mathbf{H}_\textrm{out}^\top)
\end{aligned}
\end{equation}
The MLP architecture follows \citet{kim2019cpcfg}, with details in the appendix.
This factorized parameterization, shown in Figure~\ref{fig:emit},
reduces the total parameters to  $O(h^2 + h|\mcZ| + h|\mcX|)$.

Note that parameter computation is independent of inference
and can be cached completely as the emission and transition matrices, $\bA$ and $\bO$, at test-time.
For the training algorithm, shown in Algorithm~\ref{fig:algo}, we compute $\bA$ and $\bO$ once per batch
while RNNs and similar models recompute emissions every token.

\begin{algorithm}[t]
\begin{algorithmic}
\State{Given: block structure and model parameters}
    \State{Sample block-wise dropout mask $\bb$}
    \State{Compute $\bA,\bO$ ignoring $b_z = 0$}
    \ForAll{examples $\bx$ in batch}
        \State{Compute $\log p(\bx; \bA, \bO)$}
        \State{Compute grad wrt parameters of $\log p(\bx)$}
    \EndFor
    \State{Update model parameters $\mathbf{E_z, E_x}$ and MLP}
\end{algorithmic}
\caption{
\label{fig:algo}
HMM Training (a single batch)
}
\end{algorithm}

\vspace{0.2cm}

\noindent
\textbf{Dropout as State Reduction}
Finally, to encourage full use of the large state space,
we introduce dropout that prevents the model from favoring specific states. 
We propose a form of HMM state dropout that removes states from use entirely
at each batch,
which also has the added benefit of speeding up inference.

\begin{figure}[!t]
\begin{center}
\input{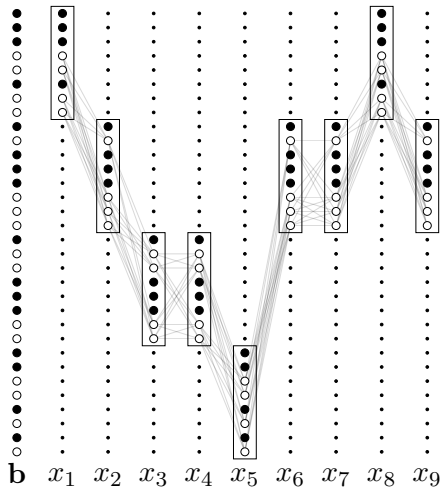}
\end{center}
\caption{
\label{fig:trellis}
The computation of $p(\bx)$ is greatly reduced by blocked emissions and state dropout.
In the above trellis, each row corresponds to a latent state and each column after 
the first to a timestep.
Each edge between nodes corresponds to a nonzero transition probability.
Blocked emissions result in a small subset of all states emitting a given word,
as shown by the rectangles. 
State dropout (leftmost column) allows us to further reduce the number of states we consider,
halving the number of (white) states that have nonzero probability in each rectangle. In experiments, the number of possible transitions may be as large as $2^{30}$ while the max number of non-zero transitions is $2^{16}$.
}
\end{figure}

State dropout acts on each emission block $\mathbf{O}_1, \ldots, \mathbf{O}_M$ independently.
For each block, we sample a binary dropout mask by sampling
$ \lambda k$ dropped row indices uniformly without replacement,
where $\lambda$ is the dropout rate.
We concatenate these into a global vector $\mathbf{b}\in\set{0,1}^{|\mcZ|}$, which, along with the previous constraints, 
ensures,
\begin{equation}
\label{eqn:state_dropout}
\begin{aligned}
p(z_t \mid z_{t-1}) &\propto b_{z_t}A_{z_{t-1}z_t}\\
p(x_t \mid z_t) &\propto b_{z_t}1(z \in \mcZ_{x_t})O_{z_t x_t}
\end{aligned}
\end{equation}
An example of the HMM lattice after state dropout is show in Figure~\ref{fig:trellis}.

In addition to accuracy improvements, state dropout gives a large practical speed up for both parameter computation and inference.
For $\lambda=0.5$ we get a $4\times$ speed improvement for both,
due to the reduction in possible transitions.
This structured dropout is also easy to exploit on GPU,
as it maintains block structure.

\section{Experimental Setup}
\label{sec:experiments}

\noindent\textbf{Emission Blocks}
The model requires partitioning token types into blocks $\mcX_m$. 
While there are many partitioning methods, a natural choice
is Brown clusters \citep{brown1992,liang2005brown} which are also based on HMMs.
Brown clusters are obtained by assigning every token type in $\mcX$ a state in an HMM,
then merging states until a desired number of partitions $M$ is reached.
We construct the Brown clusters on the training portions of the datasets
and assume the vocabulary remains identical at test time (with OOV words mapped to unk).
We include more background on Brown Clusters in the appendix.

\noindent\textbf{State Dropout}
We use a dropout rate of $\lambda = 0.5$ at training time. 
For each block of size $|\mcX_m|$, we sample $\lambda|\mcX_m|$ states to use in that block each batch.
We draw states from each block from a multivariate hypergeometric distribution
using the Gumbel Top-k trick for sampling without replacement \citep{vieira2014gumbel}.
At test time we do not use state dropout.

\noindent \textbf{Datasets}
We evaluate on the \textsc{Penn Treebank} \citep{ptb} (929k train tokens, 10k vocab)
and \textsc{Wikitext2} \citep{wikitext} (2M train tokens, 33k vocab) datasets.
For \textsc{Penn Treebank} we use the preprocessing from \citet{mikolov-2011},
which lowercases all words and substitutes OOV words with unks. 
We insert EOS tokens after each sentence.
For \textsc{Wikitext2} casing is preserved, and all OOV words are unked.
We insert EOS tokens after each paragraph.
In both datasets OOV words were included in the perplexity (as unks),
and EOS was included in the perplexity as well \citep{merity2017awdlstm}.

\noindent \textbf{Baselines}
Baselines include both state-of-the-art language models 
and other alternative LM styles. 
These include AWD-LSTM \citep{merity2017awdlstm};
a 900-state scalar HMM and HMM+RNN extension,
which discards the HMM assumptions \citep{buys2018hmm};
a traditional Kneser-Ney 5-gram model \citep{mikolov2012rnn,kenlm},
a 256 dimension feedforward neural model,
and a 2-layer 256 dimension LSTM.

We compare these with our approach: the very large neural HMM (VL-HMM).
Unless otherwise noted, our model has $|\mcZ|=2^{15}$ total states 
but only considers $k=256$ states at every timestep at
test time with $M=128$ groups.\footnote{
The 256 dim FF, LSTM, and VL-HMM in particular
have comparable computational complexity: $O(256^2 T)$.
}
The state and word embeddings as well as the MLP have a hidden dimension of 256.
We train with a state dropout rate of $\lambda=0.5$.
See the appendix for all hyperparameters.

\begin{table}[!t]
\centering
\begin{tabular}{lrrr}
\toprule
Model & Param & Val  & Test \\
\midrule
\textsc{Penn Treebank}\\
\midrule
KN 5-gram              & 2M   & - & 141.2\\
AWD-LSTM               & 24M  & 60.0 & 57.3\\
256 FF 5-gram          & 2.9M & 159.9      & 152.0  \\
2x256 dim LSTM         & 3.6M & 93.6       & 88.8   \\
HMM+RNN                & 10M  & 142.3 & -\\
HMM $|\mcZ|=900$       & 10M  & 284.6 & -\\
VL-HMM $|\mcZ|=2^{15}$ & 11.4M & 125.0      & 116.0  \\
\midrule
\textsc{WikiText}\\
\midrule
KN 5-gram               & 5.7M    & 248.7 & 234.3\\
AWD-LSTM                & 33M     & 68.6  & 65.8\\
256 FF 5-gram           & 8.8M    & 210.9 & 195.0\\
2x256  LSTM             & 9.6M    & 124.5 & 117.5\\
VL-HMM $|\mcZ|=2^{15}$  & 17.3M   & 166.6  & 158.2\\
\bottomrule
\end{tabular}
\caption{\label{tbl:ppl}
Perplexities on \textsc{PTB / Wikitext-2}.
The HMM+RNN and HMM of \citet{buys2018hmm} reported validation perplexity only for \textsc{PTB}.
}
\end{table}

\section{Results}
Table \ref{tbl:ppl} gives the main results.
On \textsc{PTB}, the VL-HMM is able to achieve 125.0 perplexity on the valid set,
outperforming a FF baseline (159.9)
and vastly outperforming the 900-state HMM from \citet{buys2018hmm} (284.6).\footnote{
\citet{buys2018hmm} only report validation perplexity
for the HMM and HMM+RNN models, so we compare accordingly.}
The VL-HMM also outperforms the HMM+RNN extension of \citet{buys2018hmm} (142.3).
These results indicate that HMMs are a much stronger model
on this benchmark than previously claimed.
However, the VL-HMM is still outperformed by LSTMs which have been extensively studied for this task.
This trend persists in \textsc{Wikitext-2},
with the VL-HMM outperforming the FF model but underperforming an LSTM.

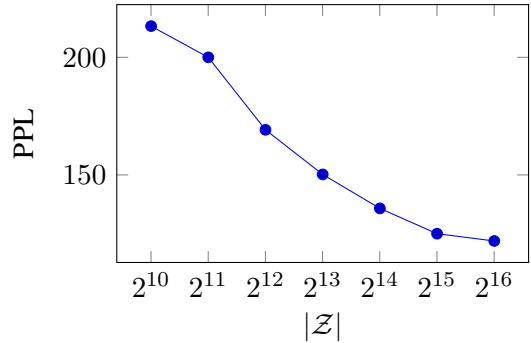
\begin{figure}[!t]
\centering
\begin{tikzpicture}
\begin{axis}[
    xlabel=$|\mcZ|$,
    ylabel=PPL,
    xmode=log,
    log basis x={2},
    xtick={},
    width=7cm,
    height=5cm
]
  \addplot plot coordinates {
    (1024,  213.25)
    (2048,  199.98)
    (4096,  169.18)
    (8192,  150.22)
    (16384, 135.79)
    (32768, 125.02)
    (65536, 121.93)
};
\end{axis}
\end{tikzpicture}
\caption{\label{tbl:states-ablation}
Perplexity on \textsc{PTB} by state size $|\mcZ|$ ($\lambda =0.5$ and $M=128$).
}
\end{figure}

Figure~\ref{tbl:states-ablation} examines the effect of state size:
We find that performance continuously improves significantly as we grow to $2^{16}$ states,
justifying the large state space.
The marginal improvement does lower as the number of states increases,
implying that the current approach may have limitations in scaling to even larger state spaces.

Table~\ref{tbl:dropout-param-ablation} considers other ablations:
Although neural and scalar parameterizations reach similar training perplexity,
the neural model generalizes better on validation
with almost 100x fewer model parameters.
We find that state dropout results in both an improvement in perplexity and
a large improvement in computational speed.
See the appendix for emission sparsity constraint ablations,
as well as experiments on further reducing the number of parameters.

\begin{table}[!t]
\centering
\begin{tabular}{lrrrr}
\toprule
Model                & Param & Train  & Val  &  Time \\
\midrule
VL-HMM ($2^{14}$)    & 7.2M & 115    & 134  & 40\\
\quad - neural param & 423M & 119    & 169  & 14\\
\quad - state dropout      & 7.2M & 88     & 157  & 100\\
\bottomrule
\end{tabular}
\caption{\label{tbl:dropout-param-ablation}
Ablations on \textsc{PTB} ($\lambda =0.5$ and $M=128$) with a smaller model $|\mcZ| = 2^{14}$. 
Time is ms per eval batch (Run on RTX 2080).
Ablations were performed independently, removing a single component per row. Removing the neural parameterization results in a scalar parameterization.
} 
\end{table}

\section{Conclusion}
This work demonstrates methods for effectively scaling HMMs to large state spaces on parallel 
hardware, and shows that this approach results in accuracy gains compared to other HMM models.
In order to scale, we introduce three techniques:
a blocked emission constraint, a neural parameterization,
and state dropout, which lead to an HMM that outperforms n-gram models and prior HMMs.
Once scaled up to take advantage of modern hardware,
very large HMMs demonstrate meaningful improvements over smaller HMMs.
HMMs are a useful class of probabilistic models with different inductive biases, 
performance characteristics, and conditional independence structure than RNNs. 
Future work includes using these approaches to induce model structure, develop accurate models 
with better interpretability, and to apply these approaches in lower data regimes.

\subsection*{Acknowledgements}
We are grateful to the anonymous reviewers, as well as
Yuntian Deng, Daniel Fried, Yao Fu, Yoon Kim, Victor Sanh, Sam Wiseman, and Jiawei Zhou
for the valuable feedback, insightful conversations, and suggestions.
 This work is supported by CAREER 2037519 and NSF III 1901030.

\bibliographystyle{acl_natbib}
\bibliography{anthology}

\appendix

\section{Appendices}

\subsection{Brown Clustering}
Brown clustering is an agglomerative clustering approach \citep{brown1992,liang2005brown}
that assigns every token type a single cluster.
The Brown clustering model aims to find an HMM that maximizes the likelihood of
an observed corpora under the constraint that every token type can only be emit
by a single latent class.
The cluster for the word is given by the latent class that emits that token type.

Clusters are initialized by assigning every token type a unique latent state in an HMM.
States are then merged iteratively until a desired number $M$ is reached.
\citet{liang2005brown} propose an algorithm that chooses a pair of states
to merge at every iteration based on state bigram statistics within a window.

\subsection{Hyperparameters}
\label{sec:hyperparams}

For \textsc{Penn Treebank} and \textsc{Wikitext-2}, we trained the following baselines:
a two layer FF 256-dim 5-gram model and a two layer 256-dim LSTM.
The FF model is given by the following:
\begin{equation}
\begin{aligned}
p(w_t \mid \bw_{<t})
&= W_x\textrm{ReLU}(W_h\mathbf{E}_w(\bw_{t-4:t-1}))
\end{aligned}
\end{equation}
where $\mathbf{E}_w$ gives the word embeddings,
$W_h\in\R^{h\times 4h}$, and
$W_x\in\mathbb{R}^{|\mcX|\times h}$ is weight-tied to the word embeddings.
The LSTM model is given by:
\begin{equation}
p(w_t \mid \bw_{<t})
= W_x \textrm{LSTM}(\mathbf{E}_w(\bw_{<t}))
\end{equation}
with a 2-layer LSTM
that has weight-tied $W_x$ and $\mathbf{E}_w$.

For the (5-gram) FF model we use a batch size of 128 and a bptt length of 64,
as we found the model needed a larger batch size to achieve decent performance.
For the LSTM, we use a batch size of 16 and a BPTT length of 32.
For both baseline models we use AdamW \citep{adamw} with a learning rate of 1e-3 and a dropout rate of 0.3 on the activations in the model.
Both models use a hidden dimension of $h=256$ throughout.
These same hyperparameters were applied on both \textsc{Penn Treebank} and \textsc{Wikitext-2}.

For the HMMs we use a batch size of 16 and a BPTT length of 32.
We use state dropout with rate $\lambda = 0.5$.
We reset the state distribution to $p(z_1 \mid z_0)$ after encountering the EOS symbol.
We use AdamW \citep{adamw} with a learning rate of 1e-2 for \textsc{Penn Treebank},
and a learning rate of 1e-3 for \textsc{Wikitext-2}.

All weights are initialized with the Kaiming uniform initialization.
The FF model was trained for 100 epochs, while all other models were trained for 50.
Validation likelihood was checked 4 times per epoch, and
learning rates were decayed by a factor of 4 if the validation performance
did not improve after 8 consecutive checks.

Hyperparameter search was performed manually, using the best
validation perplexity achieved in a run.
Bounds:
\begin{enumerate}
\item Learning rate $\in \set{0.0001, 0.001, 0.01, 0.1}$
\item Dropout $\lambda \in \set{0, 0.25, 0.5, 0.75}$
\item Hidden dimension $h \in \set{128,256,512}$
\item Batch size $\in \set{16, 32, 64, 128}$
\end{enumerate}

Experiments were run on RTX 2080 GPUs.

On \textsc{PTB} the FF model takes 3s per epoch, the LSTM 23s,
and the VLHMM $2^{15}$ 433s.
The inference for VLHMM was not heavily optimized,
and uses a kernel produced by TVM \citep{tvm} for computing
gradients through marginal inference.

\subsection{HMM Parameterization}
Let $\mathbf{E},\mathbf{D}\in\mathbb{R}^{v \times h}$ be an
embedding matrix and a matrix of the same size,
where $v$ is the size of the vocab and $h$ the hidden dimension.
We use the following residual network as our MLP:
\begin{equation}
\label{eqn:res}
\begin{aligned}
f_i(\mathbf{E}) &= g_i(\textrm{ReLU}(\mathbf{E}W_{i1}))\\
g_i(\mathbf{D}) &= \textrm{LayerNorm}(\textrm{ReLU}(\mathbf{D}W_{i2}) + \mathbf{D})
\end{aligned}
\end{equation}
with $i \in \set{\textrm{out},\textrm{in},\textrm{emit}}$,
$W_{i1},W_{i2} \in \mathbb{R}^{h \times h}$.
The state embeddings are then obtained by
\begin{equation}
\begin{aligned}
\mathbf{H}_\textrm{out} &= f_\textrm{out}(\mathbf{E}_z)\\
\mathbf{H}_\textrm{in} &= f_\textrm{in}(\mathbf{E}_z)\\
\mathbf{H}_\textrm{emit} &= f_\textrm{emit}(\mathbf{E}_z)
\end{aligned}
\end{equation}

In order to reduce the number of parameters further, we experiment with factored state embeddings. We factor the state embeddings into a composition of smaller steate embeddings ($\mathbf{E}_z'\in\R^{|\mcZ|\times h/2}$) as well as block embeddings ($\mathbf{E}_m\in\R^{|\mcZ|\times h/2}$), which are shared across all states within the same emission block, i.e. all $z \in \mcZ_x$ share a block embedding.
To compose these embeddings, we introduce new residual networks $f_j,j\in\set{o,i,e}$ similar to the above, yielding
\begin{equation}
\begin{aligned}
\mathbf{H}_\textrm{out} &= f_\textrm{out}(f_o([\mathbf{E}_m,\mathbf{E}_z']))\\
\mathbf{H}_\textrm{in} &= f_\textrm{in}(f_i([\mathbf{E}_m,\mathbf{E}_z']))\\
\mathbf{H}_\textrm{emit} &= f_\textrm{emit}(f_e([\mathbf{E}_m,\mathbf{E}_z']))
\end{aligned}
\end{equation}
We ablate the factored state embeddings in Sec.~\ref{sec:fac_ablation}.

\begin{table}[t]
\centering
\begin{tabular}{lllll}
\toprule
Constraint & $|\mcZ|$ & $k$ & $M$ & Val PPL\\
\midrule
Brown & 16384 & 512 & 32  & 137\\
Brown & 16384 & 256 & 64  & 138\\
Brown & 16384 & 128 & 128 & 134\\
Brown & 16384 & 64  & 256 & 136\\
\midrule
None  & 1024 & - & - & 180\\
Brown & 1024 & 256 & 4 & 182\\
Brown & 1024 & 128 & 8 & 194\\
\midrule
Uniform    & 8192    & 128    & -   & 150\\
Brown      & 8192    & 128    & 64  & 142\\
Uniform    & 16384   & 128    & -   & 146\\
Brown      & 16384   & 128    & 128 & 136\\
\bottomrule
\end{tabular}
\caption{\label{tbl:constraint-ablation}
Emission constraint ablations on \textsc{Penn Treebank}.
$|\mcZ|$ is the size of the hidden space,
$k$ is the size number of hidden states in each block,
and $M$ is the number of blocks.
}
\end{table}

\subsection{Emission Constraint Ablation}
Table~\ref{tbl:constraint-ablation} shows the results from 
emission constraint ablations.
With a VL-HMM that has $|\mcZ|=2^{14}$ states,
the model is insensitive to the number of blocks $M$ explorable given computational constraints.
However, with fewer states $|\mcZ|=2^{10}$ we are able to explore a lower number of blocks.
With $M=4$ blocks, the block-sparse HMM matches an unconstrained HMM
with the same number of states.
When $M=8$, the block-sparse model underperforms,
implying there may be room for improvement with the larger
HMMs that use $M > 8$ blocks.

We additionally compare the blocks induced by Brown clustering with a uniform
constraint that samples subsets of states of size $n$
independently and uniformly from $\mcZ$.
This does not admit a partitioning, which makes it difficult to apply state dropout.
We therefore zero out half of the columns of the transition matrix randomly
before normalization.
In the bottom of Table~\ref{tbl:constraint-ablation},
we find that models with uniform constraints
are consistently outperformed by models with Brown cluster constraints
as measured by validation perplexity.
The models with uniform constraints also have poor validation performance
despite better training performance, a symptom of overfitting.

These ablations demonstrate that the constraints based on
Brown clusters used in this work may not be optimal,
motivating future work that learns sparsity structure.

\subsection{Factored State Representation Ablation}
\label{sec:fac_ablation}
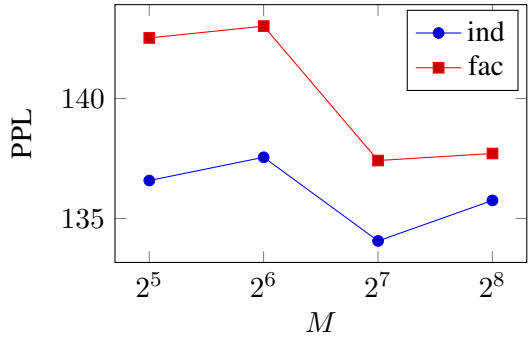
\begin{figure}[t]
\centering
\begin{tikzpicture}
\begin{axis}[
    xlabel=$M$,
    ylabel=PPL,
    xmode=log,
    log basis x={2},
    xtick={},
    width=7cm,
    height=5cm
]
  \addplot plot coordinates {
    (32,  136.58)
    (64,  137.55)
    (128, 134.06)
    (256, 135.75)
  };
  \addlegendentry{ind}
  \addplot plot coordinates {
    (32,  142.528)
    (64,  143.02)
    (128, 137.415)
    (256, 137.708)
  };
  \addlegendentry{fac}
\end{axis}
\end{tikzpicture}
\caption{\label{fig:fac-ablation}
Perplexity on \textsc{PTB} by number of blocks $M$ ($\lambda =0.5$ and $|\mcZ|=2^{14}$).
The independent embeddings (ind) represent state embeddings by directly parameterizing $\mathbf{E}_z$, while the factored embeddings (fac) compose a smaller state embeddings matrix with block embeddings.
}
\end{figure}

We examine the effect of factoring state representations into block embeddings
and independent state embeddings.
The results of the factored state ablation are in Figure~\ref{fig:fac-ablation}.
We find that the performance of independent state embeddings with
is similar to a model with factored embeddings,
but performs slightly worse in perplexity.

In Table~\ref{tbl:dropout-param-ablation-repeat} we see that although the factored state
embeddings reduce the total number of parameters,
the computation time and perplexity both get worse.

\subsection{Computational Considerations}

\begin{table}[t]
\centering
\begin{tabular}{llrrr}
\toprule
Model                & Param & Train  & Val & Time \\
\midrule
VL-HMM ($2^{14}$)    & 7.2M & 115    & 134 & 40\\
\quad - neural param & 423M & 119    & 169 & 14\\
\quad - dropout      & 7.2M & 88     & 157 & 100\\
\quad + block emb    & 5.6M & 122    & 136 & 48\\
\bottomrule
\end{tabular}
\caption{\label{tbl:dropout-param-ablation-repeat}
Ablations on \textsc{PTB} ($\lambda =0.5$ and $M=128$).
Param is the number of parameters, while train and val give the corresponding perplexities.
Time is ms per eval batch (Run on RTX 2080).
}
\end{table}

We reproduce the technique ablation table in
Table~\ref{tbl:dropout-param-ablation-repeat} for reference.
As we remove neural components, 
the number of parameters increases but the time of the
forward pass decreases.
This is because generating parameters from a neural network
takes strictly more time than having those parameters available.

When block embeddings are removed and the full state representations are
directly parameterized,
the model is faster due to not needing to recompute the full state representations.
This contrast is even more pronounced when removing neural components altogether and using a scalar parameterization, with an almost 3x speedup. This is because the distributional parameters do not need to be regenerated by a neural network if they are parameterized directly.
    
\end{document}